\pdfoutput=1

\documentclass[11pt]{article}

\usepackage[svgnames]{xcolor} 
\usepackage[final]{emnlp2023}
\usepackage{subcaption}

\usepackage{times}
\usepackage{todonotes}
\usepackage{latexsym}
\usepackage{multirow}
\usepackage[T1]{fontenc}

\usepackage[utf8]{inputenc}

\usepackage{microtype}

\usepackage{inconsolata}

\usepackage{graphicx}
\usepackage{amsmath}
\usepackage{amssymb}
\newcommand{\highlight}[3][black]{{\fboxsep3pt\colorbox{#2}{\color{#1} #3}}}

\title{Parameter-efficient Adaptation of Multilingual Multimodal Models for Low-resource ASR}

\author{Abhishek Gupta$^*\quad$ Amruta Parulekar$^*\quad$ Sameep Chattopadhyay$^*\quad$ Preethi Jyothi 
\\
 Indian Institute of Technology Bombay, Mumbai, India \\
 \small{ \texttt{\{abhishekumgupta,amrutaparulekar.iitb,sameep.ch.2002\}@gmail.com, pjyothi@cse.iitb.ac.in}}
 }

\begin{document}
\maketitle

\begin{abstract}
\vskip -0.07in
Automatic speech recognition (ASR) for low-resource languages remains a challenge due to the scarcity of labeled training data. Parameter-efficient fine-tuning and text-only adaptation are two popular methods that have been used to address such low-resource settings. In this work, we investigate how these techniques can be effectively combined using a multilingual multimodal model like SeamlessM4T. Multimodal models are able to leverage unlabeled text via text-only adaptation with further parameter-efficient ASR fine-tuning, thus boosting ASR performance. We also show cross-lingual transfer from a high-resource language, achieving up to a relative $17\%$ WER reduction over a baseline in a zero-shot setting without any labeled speech.
\vskip -0.6in
\unskip
\end{abstract}
\begin{figure*}[t] 
  \includegraphics[clip,width=16cm]{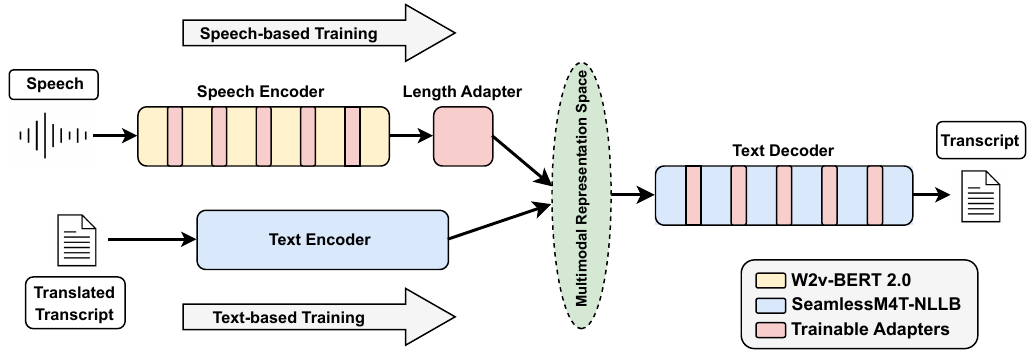} 
  \vspace {-6pt}
  \caption{\textbf{Parameter-efficient Adaptations for SeamlessM4T:} A multimodal ASR model such as SeamlessM4T can be fine-tuned in a parameter-efficient manner through either speech-based adaptations or text-only adaptation. }
  \label{fig:adapterS}
  \vspace {-6pt}
\end{figure*}
 \maketitle 
 
 \normalsize \setcounter{footnote}{0} 
 \def\thefootnote{*}\footnotetext{These authors contributed equally to this work.}
\vspace{-12pt}
\section{Introduction}
\vspace {-3pt}
Across the languages of the world, the automation of various speech and text tasks has led to the creation of massive multilingual datasets such as Multilingual LibriSpeech~\cite{Pratap_2020}, that contain speech, text, and other metadata for a number of different languages. This large-scale collection has catalyzed the emergence of large multilingual automatic speech recognition (ASR) models~\cite{yadav-sitaram-2022-survey}, which utilize the structural similarities between different languages to learn language-invariant features and boost accuracy. Subsequently, multimodal multilingual models, such as M3P \cite{ni2021m3plearninguniversalrepresentations}, that bridge the gap between speech and text using joint representation spaces, have also emerged. These models are trained using large amounts of multilingual speech and text data.

However, less-spoken languages, especially those from developing countries, do not have such large data corpora available~\cite{magueresse2020lowresourcelanguagesreviewpast}, thus hurting model performance for extremely low-resource languages~\cite{chang2023multilingualitycurselanguagemodeling}. 
Thus, creating targeted models for severely low-resource languages has become crucial. One efficient way to do this is by adapting existing models to the target language using limited amounts of labeled data. Such adaptation has to be done carefully so as to not overfit to the target language characteristics. 

Parameter-efficient fine-tuning (PEFT) \cite{han2024parameterefficientfinetuninglargemodels} techniques have gained wide acceptance where only relevant parts of a model are identified and fine-tuned for a specific downstream task. 
Text-only adaptation is another sub-area that is gaining popularity for low-resource ASR~\cite{Bataev2023TextonlyDA,vuong-etal-2023-adabert}. 
Multimodal models have training pathways for both speech and text data, offering a good framework to combine both approaches. Multilingual models, on the other hand, allow for cross-lingual transfer \cite{khare21_interspeech}, i.e., using a higher resource language to improve performance on a lower resource language. 

In this work, we have leveraged the multimodal nature of Meta's SeamlessM4T \cite{seamlessm4tmassivelymultilingual} to explore the benefits of speech-based adapter fine-tuning and text-only adaptation. These techniques have been used both in isolation and in combination to identify the best strategy to improve low-resource ASR for a number of Indic languages. We have also exploited the multilingual nature of the model to use higher-resource languages to improve low-resource ASR. Thus, our main contributions include: (a) identifying how to combine speech-based parameter-efficient fine-tuning and text-only adaptation to boost low-resource ASR, (b) identifying a cross-lingual transfer technique that can give more than $17\%$ relative reduction in WER for a low-resource language without using any speech of that language, (c) the use of small amounts of available data to boost the performance of  SeamlessM4T \cite{seamlessm4tmassivelymultilingual} on six Indic languages, Bengali, Gujarati, Kannada, Maithili, Malayalam and Odia.

\section{Related Work}
\label{sec:rel works}
One of the key challenges in current ASR  research is enabling systems to handle multilingual inputs \cite{yadav-sitaram-2022-survey,kannan} while minimizing resource requirements in terms of training, inference, and storage costs. 
Currently, the most popular paradigm using multilingual models are to initially pre-train the models in a self-supervised manner on a large multilingual dataset \cite{Babu2021XLSRSC} before being fine-tuned on a set of target languages \cite{Toshniwal/ICASSP.2018.8461972, Bai-inproceedings}.  A general way of performing such model fine-tuning is by updating all the weights or some specific model components while training. These kinds of methods are parameter inefficient and often cause catastrophic forgetting \cite{Kessler2021Continualwav2vec2AA}, for all non-target languages. Also, training and storage costs for such methods increase linearly with both the model size and the number of languages.  
\vspace{3pt} 

To mitigate these limitations, recent literature on NLP has introduced several parameter-efficient fine-tuning methods \cite{xu2023parameterefficientfinetuningmethodspretrained,tomanek-etal-2021-residual,hu2021loralowrankadaptationlarge}, often involving trainable modules called adapters \cite{PETNLP}, whose weights are updated while freezing the original backbone. Significant efforts are being made to develop better adapter architectures and efficient training methods \cite{10.5555/3618408.3620103}  to utilize contrastive learning \cite{zhang/3600270.3601846} and meta-learning \cite{houinproceedings}. These modules can also be used to adapt multilingual ASR models for a low-resource setting, with Simadapter \cite{hou/TASLP.2021.3138674} being one of the first models to utilize adapters to leverage cross-lingual features. 

In the context of speech recognition, a low-resource setting could refer to any scenario with insufficient training data. This includes challenges such as recognizing atypical speech \cite{tomanek-etal-2021-residual} or processing less commonly spoken languages. A recent work \cite{mainzinger-levow-2024-fine} demonstrated the benefits of using adapters for very low-resource languages with less than five hours of training data. For the low-resource situation, task- or language-specific adapter modules showcase superior performance \cite{HU2024103037} compared to fine-tuning the model components, but even such approaches are constrained by inherent limitations of the base model.

Over the past few years, considerable effort has gone into developing multilingual ASR foundational models with more generalizable features. These models offer a stronger starting point for low-resource adaptations and enable the use of cross-lingual transfer learning. The exponential growth in computing power has led to the creation of increasingly large language models, which are now used for a wide range of tasks, including as backbones for multimodal ASR models \cite{rubenstein2023audiopalmlargelanguagemodel,zhang2023speechgptempoweringlargelanguage,chang2023multilingualitycurselanguagemodeling}. For such models, the foundational backbone is expanded using audio tokens generated using techniques like wav2vec~\cite{schneider2019wav2vecunsupervisedpretrainingspeech} and Hubert~\cite{hubert/TASLP.2021.3122291} in order to learn a joint representation in a multimodal space; the token vocabulary is expanded to encompass both text and audio. Note that models with joint multimodal representations are not only useful for ASR but can also be integrated with a vocoder for TTS or conversational chatbots~\cite{zhang2023speechgptempoweringlargelanguage}.

Multimodal models can be trained with joint text-audio tasks through self-supervision with masked language modeling and denoising objectives; further fine-tuning is often done with ASR and speech-to-text or speech-to-speech translation tasks. One of the most recent examples of such a multilingual multimodal model has been SeamlessM4T~\cite{seamlessm4tmassivelymultilingual} by Meta AI, which is built upon the NLLB \cite{team2022language} backbone and can process speech and text inputs from nearly 100 languages. An implicit advantage of using such multimodal models for low-resource ASR is the ability to benefit from text-only learning for shared parameters. In most cases, there is significantly more text data available than speech data. Thus, the capability to leverage text-only adaptation for ASR models can be highly advantageous in these scenarios.

While there is a lot of prior work in the domain of text-only adaptation for ASR~\cite{vuong-etal-2023-adabert,Bataev2023TextonlyDA,10389682,mittal2023insitu}, and there has been some work on a comparative analysis of various fine-tuning strategies for low-resource ASR \cite{Liu-2024}, to the best of our knowledge, our work is the first to explore them for multilingual multimodal models.


\section{Methodology}
In this work, we leverage a combination of parameter-efficient adaptation, unlabeled textual data, and minimal amounts of transcribed speech to improve ASR performance in low-resource languages using multilingual multimodal models. Figure~\ref{fig:adapterS} demonstrates the overall workflow of our proposed pipeline. 

\subsection{Multimodal base model: SeamlessM4T}
We use SeamlessM4T~\cite{seamlessm4tmassivelymultilingual} as our base model for all our experiments. SeamlessM4T, i.e., Massively Multilingual \& Multimodal Machine Translation, is a versatile end-to-end model that provides support for multiple tasks, including speech-to-speech translation, speech-to-text translation, text-to-speech translation, text-to-text translation, and automatic speech recognition for up to 100 languages. The model has been trained using over a million hours of unlabeled speech in a self-supervised manner, along with more than 400K hours of human and machine-labeled audio. It supports 96 different languages for input speech and text, as well as output text, and can generate speech in 35 languages. 

The SeamlessM4T model architecture is inspired by UnitY~\cite{inaguma-etal-2023-unity}, a two-pass modeling framework that, unlike cascaded models, can be jointly optimized.  The text encoder and decoder models of SeamlessM4T are initialized by the NLLB model~\cite{nllbteam2022languageleftbehindscaling}, a text-to-text translation model. To process speech inputs, the model employs the Wav2Vec-BERT 2.0 speech encoder, which is an enhancement over the original model proposed by ~\citet{Chung2021w2vBERTCC} with additional codebooks. The model also includes a modality adapter~\cite{zhao22g_interspeech}, referred to as the \textbf{length adapter}, to align the speech modality with text, projecting it to a unified representation space. Lastly, the model uses a text-to-unit (T2U) component for speech generation that produces discrete speech units from the text output. These units are then transformed into audio waveforms using a multilingual HiFi-GAN unit vocoder~\cite{HIFI-GAN/3495724.3497152}. There are multiple variants of the SeamlessM4T model; we have used SeamlessM4T-medium with a total of 1.2 Billion parameters.

Although the entire model comprises multiple components, our analysis focuses primarily on applying SeamlessM4T for multilingual ASR. The ASR pipeline of SeamlessM4T consists of the speech encoder (311M parameters), the length adapter (46M parameters), and the text decoder (201M parameters). Next, we will elaborate on parameter-efficient fine-tuning of SeamlessM4T (Section~\ref{sec:peft}) and how we can use text-only adaptation within such a multimodal model (Section~\ref{sec:text-only}).




\subsection{Parameter-efficient Fine-tuning}
\label{sec:peft}

The ASR components of SeamlessM4T amount to more than 500M parameters. Full fine-tuning of these components using limited amounts of labeled data for low-resource languages may result in overfitting and degradation of ASR performance. To alleviate these challenges, parameter-efficient fine-tuning paradigms like the \emph{adapter framework}~\cite{PETNLP} are very popular, especially for natural language processing tasks. Adapters have also found success in low-resource ASR tasks such as accent adaptation~\cite{tomanek-etal-2021-residual} and cross-lingual adaptation~\cite{hou/TASLP.2021.3138674}. 
Next, we will elaborate on the structure of an existing \emph{length adapter} within SeamlessM4T and the new adapters we introduce in the encoder and decoder layers.

\subsubsection{The Length Adapter}

The length adapter in SeamlessM4T aims to bridge the gap between speech and text representations. It is inspired by the M-adapter architecture~\cite{zhang2023speechgptempoweringlargelanguage} and uses a Transformer-based module to adapt speech representations to text. By compressing the speech sequence, the length adapter generates features tailored for multilingual speech-to-text tasks by modeling both global and local dependencies within the speech.   
\vspace{3pt}

\begin{figure}[t] 
  \includegraphics[angle=270,clip,width=8cm,trim=0cm 2cm 6cm 2cm]{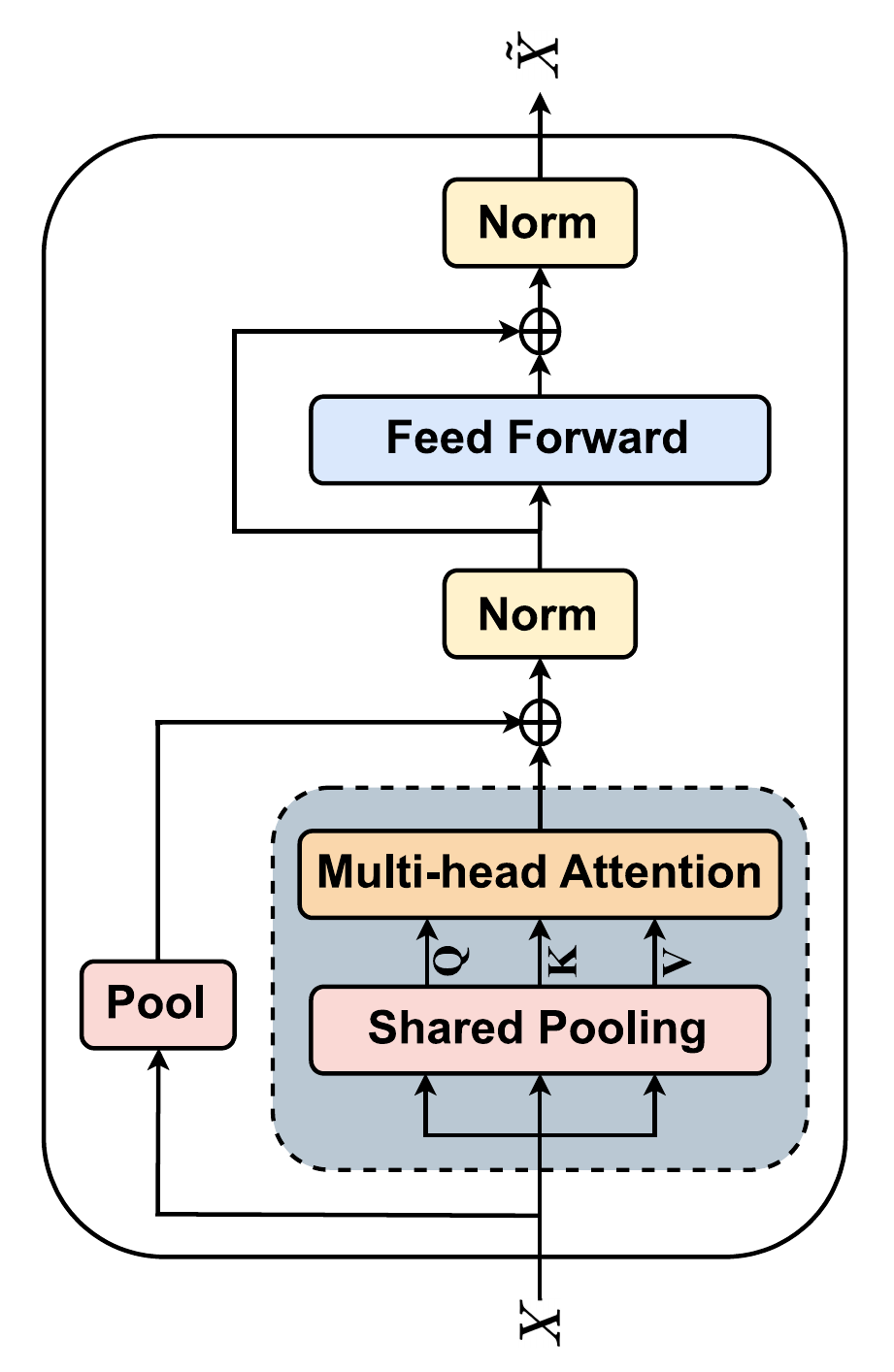}
  \caption{\textbf{SeamlessM4T Length Adapter:} Projects speech embedding \(X\) to a lower-dimensional representation \(\tilde{X}\) in the multimodal space.}
  \label{fig:len adap}
\vspace {-3pt}
\end{figure}

The main part of the original M-adapter architecture, illustrated in Figure~\ref{fig:len adap}, is the Multi-head Pooled Self-Attention (MPSA) mechanism. In the original MPSA, convolutional layers pool the input \(X\) and are further projected to the inputs of the multi-head attention module using linear transformation matrices. An additional pooling is applied in parallel to \(X\) and then added to the output of the attention module before being processed through a feedforward network. These processes together generate a lower dimensional representation of \(X\), denoted by \(\tilde{X}\) as the current layer output, addressing any length mismatches between embeddings from different modalities. Unlike the original M-adapter architecture with independent pooling modules for the multi-head attention inputs, the length adapter utilizes a shared pooling module, generating a single \(\hat{X}\) for each \(X\) to improve efficiency. 
\vspace{3pt}
\vspace{-0.3pt}
More formally, given an input sequence \( X \in \mathbb{R}^{L \times D} \), where \( L \) is the sequence length and \( D \) is the embedding dimension, the MPSA mechanism starts by applying shared pooling to the input \( X \) to obtain \( \hat{X} \in \mathbb{R}^{L' \times D} \). This pooling operation is performed using a 1D convolutional layer with kernel size \( k \), stride \( s \), and padding \( p \). Subsequently, \( \hat{X} \) is linearly projected into the query, key, and value matrices, denoted as \( Q \), \( K \), and \( V \), respectively.
\begin{align*}
\hat{X} &= \;\mathrm{SharedPooling}\left(X\right) \\
\vspace{3pt}
Q &= \;\hat{X} W^{Q}, \quad &&\text{where } Q \in \mathbb{R}^{L' \times D}, \\
K &= \;\hat{X} W^{K}, \quad &&\text{where } K \in \mathbb{R}^{L' \times D}, \\
V &=\; \hat{X} W^{V}, \quad &&\text{where } V \in \mathbb{R}^{L' \times D}.
\end{align*}
\vspace{3pt}
\noindent where the new sequence length $L'$ is given by:
\[
L' = \left\lfloor \frac{L + 2p - k}{s} \right\rfloor + 1.
\]
\vspace{3pt}


We hypothesize that the length adapter module could potentially learn prosodic characteristics of languages, such as phoneme durations, by mapping speech embeddings --- which include both segmental and suprasegmental information --- to text embeddings that contain only content information. Learning certain prosodic characteristics like durations can be particularly beneficial for extremely low-resource languages that lack sufficient data for learning fine-grained contextual and syntactical information.
\vspace{3pt}
\subsubsection{Encoder and Decoder Adapters}
\vspace{3pt}
In addition to the pre-existing length adapter (Figure \ref{fig:len adap}) in the SeamlessM4T architecture, we inserted additional trainable adapter layers within the encoder and decoder modules to adapt this multilingual model for low-resource languages. The adapter modules, following the architecture proposed in \cite{PETNLP}, initially project the original \(D_1\)-dimensional features into an intermediate space of dimension \(D_2\). A non-linearity, specifically GeLU~\cite{hendrycks2023gaussianerrorlinearunits} in our implementation, is then applied, after which the features are projected back to the original \(D_1\) dimensions. To adjust the number of parameters for these adapters, we can change the intermediate dimension \(D_2\). By decreasing the value of \(D_2\), the number of trainable parameters in the adapters is reduced accordingly.
\vspace{3pt}

In our current experimental setup, we have inserted adapters after every Conformer layer in the encoders and after every Transformer layer in the text decoder. By setting the intermediate dimension \(D_2\) to one-fourth of  \(D_1\) for all adapters, we introduce 6 million new trainable parameters each in the encoder and decoder modules.

Formally, the operations inside the $i^{\text{th}}$ speech encoder layer can be summarized as:
\vspace{-0.1cm}
\begin{align*}
    \mathbf{H} &= \mathrm{MultiHeadAttn}(\mathbf{h}^{i-1},\mathbf{h}^{i-1},\mathbf{h}^{i-1}) \nonumber \\[-2pt]
    \mathbf{C} &= \mathrm{Convolution}(\mathbf{H}) \nonumber \\[-2pt]
    \mathbf{\hat{h}^i} &= \mathrm{FFN}(\mathbf{C}) \nonumber \\[-2pt]
    \mathbf{h}^i &= \mathrm{Adapter}(\mathbf{\hat{h}}^i)
\end{align*}
Similarly, the operations inside the $i^{\text{th}}$ decoder layer can be summarized as:
\vspace{-0.1cm}
\begin{align*}
    \mathbf{D} &= \mathrm{MultiHeadAttn}(\mathbf{d}^{i-1},\mathbf{d}^{i-1},\mathbf{d}^{i-1}) \nonumber \\[-2pt]
    \mathbf{\hat{D}} &= \mathrm{MultiHeadAttn}(\mathbf{d}^{i-1},\mathbf{h}^{\ell},\mathbf{h}^{\ell}) \nonumber \\[-2pt]
    \mathbf{\hat{d}^i} &= \mathrm{FFN}(\mathbf{\hat{D}}) \nonumber \\[-2pt]
    \mathbf{d}^i &= \mathrm{Adapter}(\mathbf{\hat{d}}^i)
\end{align*}
where $\ell$ is the last encoder layer, and MultiHeadAttn(Q, K, V) is the standard multi-head attention implementation \cite{vaswani2017attention} with Q, K, and V denoting queries, keys, and values, respectively.

During our experiments, we fine-tuned the encoder adapters and length adapters on labeled ASR data, while the decoder was fine-tuned using ASR and machine translation (MT) data, thereby leveraging the text-to-text pipeline of SeamlessM4T. 



\begin{table*}[]
\centering
\resizebox{\textwidth}{!}{%
\renewcommand{\arraystretch}{1.1}
\begin{sc}
\begin{tabular}{l|l|ll|ll|ll|ll|ll|ll}
\hline
\begin{tabular}[c]{
@{}l@{}}Components\\ \end{tabular} &
  \begin{tabular}[c]{@{}l@{}}Learnable\\ \end{tabular} &
  \multicolumn{2}{c|}{Maitihli} &
  \multicolumn{2}{c|}{Malayalam} &
  \multicolumn{2}{c|}{Kannada} &
  \multicolumn{2}{c|}{Gujarati} &
  \multicolumn{2}{c|}{Odia} 
  &
  \multicolumn{2}{c}{Bengali} 
  \\ \cline{3-14}
  fine-tuned &
Parameters &
  \multicolumn{1}{l|}{WER} &
  CER &
  \multicolumn{1}{l|}{WER} &
  CER &
  \multicolumn{1}{l|}{WER} &
  CER &
  \multicolumn{1}{l|}{WER} &
  CER &
  \multicolumn{1}{l|}{WER} &
  CER 
  &
  \multicolumn{1}{l|}{WER} &
  CER 
  \\ \hline
  None & -
   &
  \multicolumn{1}{l|}{82.20} & 43.39
   &
  \multicolumn{1}{l|}{56.15} & 20.65
   &
  \multicolumn{1}{l|}{69.29} & 29.11
   &
  \multicolumn{1}{l|}{41.03} & 24.50
   &
  \multicolumn{1}{l|}{42.81} & 17.38
  &
  \multicolumn{1}{l|}{37.70} & 18.44
   \\ 
 Length adapter  & 46M
   &
  \multicolumn{1}{l|}{54.97} & 26.10
   &
  \multicolumn{1}{l|}{52.82} & 18.14
   &
  \multicolumn{1}{l|}{55.48} & 20.38
   &
  \multicolumn{1}{l|}{33.91} & 16.40
   &
  \multicolumn{1}{l|}{35.48} & 13.75
  &
  \multicolumn{1}{l|}{35.90} & 17.08
   \\ 
  Text Decoder  & 201M
   &
  \multicolumn{1}{l|}{54.56} & 26.21
   &
  \multicolumn{1}{l|}{54.04} & 19.28
   &
  \multicolumn{1}{l|}{54.3} & 20.57
   &
  \multicolumn{1}{l|}{33.62} & 17.12
   &
  \multicolumn{1}{l|}{35.14} & 13.48
  &
  \multicolumn{1}{l|}{36.14} & 17.95
   \\
 Speech Encoder & 311M
   &
  \multicolumn{1}{l|}{43.87} & 17.79
   &
  \multicolumn{1}{l|}{46.99} & 13.45
   &
  \multicolumn{1}{l|}{47.91} & 14.93
   &
  \multicolumn{1}{l|}{27.79} & 11.58
   &
  \multicolumn{1}{l|}{29.82} & 9.24
  &
  \multicolumn{1}{l|}{29.07} & 12.09
   \\
   \hline
\end{tabular}%

\end{sc}
}
\caption{\textbf{Fine-tuning a Multimodal Model:} Comparison of WER (\%) and CER (\%) after ASR fine-tuning of SeamlessM4T with 5 hours of labeled speech, without adaptations; the first row presents the pre-fine-tuning results.}
  \label{table system A-1}
\end{table*}
\begin{table*}[]
\centering
\resizebox{\textwidth}{!}{%
\renewcommand{\arraystretch}{1.1}
\begin{sc}

\begin{tabular}{l|l|ll|ll|ll|ll|ll|ll}
\hline
\begin{tabular}[c]{@{}l@{}}Text-only\end{tabular} &
  \begin{tabular}[c]{@{}l@{}}Learnable\end{tabular} &
  \multicolumn{2}{c|}{Maithili} &
  \multicolumn{2}{c|}{Malayalam} &
  \multicolumn{2}{c|}{Kannada} &
  \multicolumn{2}{c|}{Gujarati} &
  \multicolumn{2}{c|}{Odia} 
   & \multicolumn{2}{c}{Bengali} 
  \\ \cline{3-14}
 Adaptation&
  Parameters &
  \multicolumn{1}{l|}{WER} &
  CER &
  \multicolumn{1}{l|}{WER} &
  CER &
  \multicolumn{1}{l|}{WER} &
  CER &
  \multicolumn{1}{l|}{WER} &
  CER &
  \multicolumn{1}{l|}{WER} &
  CER 
   & \multicolumn{1}{l|}{WER} &
   CER
  \\ \hline
 None & -
   &
  \multicolumn{1}{l|}{82.20} & 43.39
   &
  \multicolumn{1}{l|}{56.15} & 20.65
   &
  \multicolumn{1}{l|}{69.29} & 29.11
   &
  \multicolumn{1}{l|}{41.03} & 24.50
   &
  \multicolumn{1}{l|}{42.81} & 17.38
    &\multicolumn{1}{l|}{37.70} & 18.44
   \\ 
 5hr Transcript & 6M
   &
  \multicolumn{1}{l|}{71.32} & 37.92
   &
  \multicolumn{1}{l|}{\textbf{53.96}} & \textbf{18.94}
   &
  \multicolumn{1}{l|}{70.52} & 32.54
   &
  \multicolumn{1}{l|}{35.67} & 19.19
   &
  \multicolumn{1}{l|}{38.77} & \textbf{14.84}
    &\multicolumn{1}{l|}{\textbf{35.28}} & \textbf{16.77}
   \\
 Full Transcript & 6M
   &
  \multicolumn{1}{l|}{\textbf{68.24}} & \textbf{36.84}
   &
  \multicolumn{1}{l|}{55.30} & 20.43
   &
  \multicolumn{1}{l|}{\textbf{68.13}} & \textbf{26.91}
   &
  \multicolumn{1}{l|}{\textbf{35.45}} & \textbf{18.66}
   &
  \multicolumn{1}{l|}{\textbf{38.39}} & 16.22
    &\multicolumn{1}{l|}{35.44} & 17.73
   \\
   \hline
\end{tabular}%
\end{sc}
}
\caption{\textbf{Text-only Adaptation:} Comparison of WER (\%) and CER (\%) after text-only adaptation on SeamlessM4T with Eng-X parallel text using the full dataset and 
 a 5-hour subset; the first row presents the pre-adaptation results.}
  \label{table system B-1}
  \vspace {-7pt}
\end{table*}

\subsection{Text-only Adaptation} 

\label{sec:text-only}
The text decoder in the SeamlessM4T model is shared between the ASR pipeline and the text-to-text translation pipeline, allowing it to be trained for both tasks. This shared component in multimodal models possesses the ability to transfer knowledge from one task to another, thereby simultaneously enhancing the performance of multiple tasks.
We hypothesize that we can improve the ASR performance for a target language by fine-tuning the text decoder adapters via text-to-text translation into that language. This allows us to perform a purely text-only fine-tuning of ASR models and is especially beneficial for languages where speech data is scarce. With the latest advancements in NLP, the quality of machine-translation models has greatly improved, allowing these models to be utilized to augment the existing parallel text using machine-translated text for these languages.

\vspace{3pt}
In our text-only fine-tuning experiments, we fine-tuned the decoder adapters on an English-to-target language translation task to help them learn the relevant syntactical features for the target language.

\section{Experimental Setup}

\subsection{Dataset}
\label{sec:dataset}

The \textbf{IndicVoices} dataset \cite{javed-etal-2024-indicvoices} was utilized for all our experiments. This dataset is a multilingual, multi-speaker collection of natural and spontaneous speech in 22 Indian languages. It comprises $9\%$ read speech, $74\%$ extempore speech, and $17\%$ conversational speech. Among these languages, Maithili is classified as a zero-shot language for SeamlessM4T, while Bengali is the sole high-resource Indic language. The remaining languages are categorized as low-resource languages for the model \cite{seamlessm4tmassivelymultilingual}. One of the main reasons for using this dataset is that it is among the most comprehensive open-source, multilingual speech datasets for Indic languages covering many low-resource languages and one of the few published after the release of SeamlessM4T, ensuring there is no data leakage between the evaluation sets and the SeamlessM4T training data. 



 \subsubsection{Transcribed Speech Data} 
The speech data and the corresponding transcripts from the IndicVoices dataset were used for the ASR fine-tuning experiments. The dataset, primarily consisting of extempore speech recorded under natural conditions, is characterized by a significant amount of noise and includes occasional disfluencies. For each language, 5 hours of speech were selected for the training set, sourced from an average of 336 speakers, to simulate an extremely low-resource setting. On average, each of the test and validation sets had 1 hour of speech by 68 and 206 speakers respectively. The out-of-vocabulary (OOV) rate of the test set was calculated to determine the amount of test-train domain overlap in the data. The OOV rates for Gujarati, Bengali, Kannada, Maithili, Malayalam, and Odia test sets were $39\%$, $35\%$, $58\%$, $41\%$, $53\%$, and $37\%$, respectively, averaging to an OOV of $43.87\%$ on the test sets, further demonstrating the challenging nature of the task.
  


\subsubsection{Text-only Data} 
The \textbf{IndicTrans2} \cite{gala2023indictrans} model was used to translate all the transcriptions present in the IndicVoices dataset to obtain parallel English-X text. Another set of parallel text data was created by using only the transcriptions of the 5-hour speech data in the training set for every language. For Bengali, Gujarati, Kannada, Maithili, Malayalam, and Odia, the number of tokens in the 5-hour text sets were 40k, 43k, 30k, 42k, 34k, and 34k, respectively, while those in the large text set were 785k, 118k, 297k, 834k, 398k and 503k respectively. Thus, on average, each of the larger text data sets contained 489000 tokens for every language, while each of the smaller sets contained only 37261 tokens. 

\subsubsection{Implementation Details}


The SeamlessM4T model comprises a speech encoder with 12 Conformer blocks and a text decoder with 12 Transformer blocks, with a model dimension $D_1 = 1024$. Two $D_2$ configurations were tested: $D_2 = 256$ (about 500K parameters per adapter layer, totaling 6M parameters) and $D_2 = 2048$ (matching adapter parameters with the length adapter, totaling 50M parameters). Text-only adaptation needed roughly 200 epochs of fine-tuning, while ASR fine-tuning required up to 40 epochs. All experiments were performed with a learning rate of $5\times10^{-6}$ and a batch size of 16.


\section{Experiments and Results}
\begin{table*}[t]
\centering
\resizebox{\textwidth}{!}{%

\begin{sc}
\begin{tabular}{l|c|cccccccccccc|cc}
\hline
\multirow{6}{*}{Language} &
  \multicolumn{1}{|c|}{\multirow{2}{*}{Component}} &
  \multicolumn{2}{c}{\multirow{2}{*}{None}} &
  \multicolumn{2}{c}{\multirow{2}{*}{Length}} &
  \multicolumn{2}{c}{\multirow{2}{*}{Encoder}} &
  \multicolumn{2}{c}{\multirow{2}{*}{Decoder}} &
  \multicolumn{2}{c}{\multirow{2}{*}{Len+Enc}} &
  
  \multicolumn{2}{c}{\multirow{2}{*}{Encoder}} &
  \multicolumn{2}{|c}{\multirow{2}{*}{All}} \\  
  \rule{0pt}{3ex} 
 &
  \multicolumn{1}{|c|}{Fine-tuned} &
  \multicolumn{2}{c}{} &
  \multicolumn{2}{c}{adapter} &
  \multicolumn{2}{c}{adapter} &
  \multicolumn{2}{c}{adapter} &
  \multicolumn{2}{c}{adapter} &
  \multicolumn{2}{c}{adapter (L)} &
  \multicolumn{2}{|c}{Components} \\ 

\cline{2-16}
 &
  \multicolumn{1}{|l|}{\multirow{2}{*}{Learnable}} &
  \multicolumn{2}{c}{\multirow{2}{*}{-}} &
    \multicolumn{2}{c}{\multirow{2}{*}{46 M}} &
  \multicolumn{2}{c}{\multirow{2}{*}{6 M}} &
  \multicolumn{2}{c}{\multirow{2}{*}{6 M}} &
  \multicolumn{2}{c}{\multirow{2}{*}{52 M}} &
  
  \multicolumn{2}{c}{\multirow{2}{*}{50 M}} &
  \multicolumn{2}{|c}{\multirow{2}{*}{571 M}} \\ 
  \rule{0pt}{3ex} 
 &
  \multicolumn{1}{|l|}{Parameters} &
   &
   &
   &
   &
   &
   &
   &
   &
   &
   &
   &
   &
   &
   \\ \cline{2-16}
   &
  \multicolumn{1}{|c|}{\multirow{2}{*}{System}} &
  \multirow{2}{*}{A} &
  \multirow{2}{*}{T-A} &
  \multirow{2}{*}{A} &
  \multirow{2}{*}{T-A} &
  \multirow{2}{*}{A} &
  \multirow{2}{*}{T-A} &
  \multirow{2}{*}{A} &
  \multirow{2}{*}{T-A} &
  \multirow{2}{*}{A} &
  \multirow{2}{*}{T-A} &
  \multirow{2}{*}{A} &
  \multirow{2}{*}{T-A} &
  \multirow{2}{*}{A} &
  \multirow{2}{*}{T-A} \\
 &
  \multicolumn{1}{|l|}{} &
   &
   &
   &
   &
   &
   &
   &
   &
   &
   &
   &
   &
   &
   \\ \hline
\multirow{4}{*}{Maithili} &
  \multirow{2}{*}{WER} &
  \multirow{2}{*}{82.20} &
  \multirow{2}{*}{68.24} &
  \multirow{2}{*}{54.97} &
  \multirow{2}{*}{54.74} &
  \multirow{2}{*}{52.95} &
  \multirow{2}{*}{48.14} &
  \multirow{2}{*}{63.52} &
  \multirow{2}{*}{58.39} &
  \multirow{2}{*}{47.92} &
  \multirow{2}{*}{45.98} &
  \multirow{2}{*}{\underline{46.08}} &
  \multirow{2}{*}{\textbf{\highlight{green!50}{44.60}}} &
  \multirow{2}{*}{42.58} &
  \multirow{2}{*}{46.54} \\
 &
   &
   &
   &
   &
   &
   &
   &
   &
   &
   &
   &
   &
   &
   &
   \\ \cline{2-16}
 &
  \multirow{2}{*}{CER} &
  \multirow{2}{*}{43.39} &
  \multirow{2}{*}{36.84} &
  \multirow{2}{*}{26.10} &
  \multirow{2}{*}{27.10} &
  \multirow{2}{*}{22.86} &
  \multirow{2}{*}{21.58} &
  \multirow{2}{*}{31.60} &
  \multirow{2}{*}{29.70} &
  \multirow{2}{*}{20.56} &
  \multirow{2}{*}{20.47} &
  \multirow{2}{*}{\underline{\highlight{green!50}{19.20}}} &
  \multirow{2}{*}{\textbf{19.52}} &
  \multirow{2}{*}{17.14} &
  \multirow{2}{*}{20.78}
   \\
 &
   &
   &
   &
   &
   &
   &
   &
   &
   &
   &
   &
   &
   &
   &
   \\ \hline\multirow{4}{*}{Malayalam} &
  \multirow{2}{*}{WER} &
  \multirow{2}{*}{56.15} &
  \multirow{2}{*}{55.3} &
  \multirow{2}{*}{52.82} &
  \multirow{2}{*}{52.51} &
  \multirow{2}{*}{49.71} &
  \multirow{2}{*}{50.14} &
  \multirow{2}{*}{56.03} &
  \multirow{2}{*}{53.71} &
  \multirow{2}{*}{48.22} &
  \multirow{2}{*}{48.19} &
  \multirow{2}{*}{\underline{47.81}} &
  \multirow{2}{*}{\textbf{\highlight{green!50}{47.75}}} &
  \multirow{2}{*}{47.38} &
  \multirow{2}{*}{45.9} \\
 &
   &
   &
   &
   &
   &
   &
   &
   &
   &
   &
   &
   &
   &
   &
   \\ \cline{2-16}
 &
  \multirow{2}{*}{CER} &
  \multirow{2}{*}{20.65} &
  \multirow{2}{*}{20.43} &
  \multirow{2}{*}{18.14} &
  \multirow{2}{*}{18.87} &
  \multirow{2}{*}{15.34} &
  \multirow{2}{*}{16.35} &
  \multirow{2}{*}{20.21} &
  \multirow{2}{*}{20.00} &
  \multirow{2}{*}{14.76} &
  \multirow{2}{*}{15.46} &
  \multirow{2}{*}{\underline{\highlight{green!50}{14.12}}} &
  \multirow{2}{*}{\textbf{14.92}} &
  \multirow{2}{*}{13.86} &
  \multirow{2}{*}{13.38}
   \\
 &
   &
   &
   &
   &
   &
   &
   &
   &
   &
   &
   &
   &
   &
   &
   \\ \hline
   \multirow{4}{*}{Kannada} &
  \multirow{2}{*}{WER} &
  \multirow{2}{*}{69.29} &
  \multirow{2}{*}{68.13} &
  \multirow{2}{*}{55.48} &
  \multirow{2}{*}{53.83} &
  \multirow{2}{*}{52.54} &
  \multirow{2}{*}{53.29} &
  \multirow{2}{*}{62.88} &
  \multirow{2}{*}{58.71} &
  \multirow{2}{*}{49.36} &
  \multirow{2}{*}{48.24} &
  \multirow{2}{*}{\underline{49.14}} &
  \multirow{2}{*}{\textbf{\highlight{green!50}{47.75}}} &
  \multirow{2}{*}{45.48} &
  \multirow{2}{*}{43.5} \\
 &
   &
   &
   &
   &
   &
   &
   &
   &
   &
   &
   &
   &
   &
   &
   \\ \cline{2-16}
 &
  \multirow{2}{*}{CER} &
  \multirow{2}{*}{29.11} &
  \multirow{2}{*}{26.91} &
  \multirow{2}{*}{20.38} &
  \multirow{2}{*}{20.94} &
  \multirow{2}{*}{16.95} &
  \multirow{2}{*}{18.84} &
  \multirow{2}{*}{23.76} &
  \multirow{2}{*}{23.44} &
  \multirow{2}{*}{15.63} &
  \multirow{2}{*}{16.51} &
  \multirow{2}{*}{\underline{15.26}} &
  \multirow{2}{*}{\textbf{\highlight{green!50}{14.92}}} &
  \multirow{2}{*}{14.06} &
  \multirow{2}{*}{14.18}
   \\
 &
   &
   &
   &
   &
   &
   &
   &
   &
   &
   &
   &
   &
   &
   &
   \\ \hline
\multirow{4}{*}{Gujarati} &
  \multirow{2}{*}{WER} &
  \multirow{2}{*}{41.03} &
  \multirow{2}{*}{35.45} &
  \multirow{2}{*}{33.91} &
  \multirow{2}{*}{34.41} &
  \multirow{2}{*}{29.20} &
  \multirow{2}{*}{\textbf{\highlight{green!50}{27.72}}} &
  \multirow{2}{*}{38.88} &
  \multirow{2}{*}{35.53} &
  \multirow{2}{*}{\underline{28.03}} &
  \multirow{2}{*}{{27.73}} &
  \multirow{2}{*}{28.09} &
  \multirow{2}{*}{27.90} &
  \multirow{2}{*}{25.56} &
  \multirow{2}{*}{26.31} \\
 &
   &
   &
   &
   &
   &
   &
   &
   &
   &
   &
   &
   &
   &
   &
   \\ \cline{2-16}
 &
  \multirow{2}{*}{CER} &
  \multirow{2}{*}{24.50} &
  \multirow{2}{*}{18.66} &
  \multirow{2}{*}{16.40} &
  \multirow{2}{*}{17.41} &
  \multirow{2}{*}{\underline{\highlight{green!50}{11.96}}} &
  \multirow{2}{*}{\textbf{12.05}} &
  \multirow{2}{*}{19.28} &
  \multirow{2}{*}{17.80} &
  \multirow{2}{*}{12.63} &
  \multirow{2}{*}{12.35} &
  \multirow{2}{*}{{12.00}} &
  \multirow{2}{*}{12.50} &
  \multirow{2}{*}{11.28} &
  \multirow{2}{*}{11.67}
   \\
 &
   &
   &
   &
   &
   &
   &
   &
   &
   &
   &
   &
   &
   &
   &
   \\ \hline
   \multirow{4}{*}{Odia} &
  \multirow{2}{*}{WER} &
  \multirow{2}{*}{42.81} &
  \multirow{2}{*}{38.39} &
  \multirow{2}{*}{35.48} &
  \multirow{2}{*}{34.99} &
  \multirow{2}{*}{32.03} &
  \multirow{2}{*}{32.97} &
  \multirow{2}{*}{38.55} &
  \multirow{2}{*}{36.24} &
  \multirow{2}{*}{30.09} &
  \multirow{2}{*}{31.18} &
  \multirow{2}{*}{\underline{30.04}} &
  \multirow{2}{*}{\textbf{\highlight{green!50}{28.92}}} &
  \multirow{2}{*}{30.54} &
  \multirow{2}{*}{30.17} \\
 &
   &
   &
   &
   &
   &
   &
   &
   &
   &
   &
   &
   &
   &
   &
   \\ \cline{2-16}
 &
  \multirow{2}{*}{CER} &
  \multirow{2}{*}{17.38} &
  \multirow{2}{*}{16.22} &
  \multirow{2}{*}{13.75} &
  \multirow{2}{*}{14.62} &
  \multirow{2}{*}{10.57} &
  \multirow{2}{*}{11.25} &
  \multirow{2}{*}{14.50} &
  \multirow{2}{*}{14.57} &
  \multirow{2}{*}{10.11} &
  \multirow{2}{*}{11.32} &
  \multirow{2}{*}{\underline{10.01}} &
  \multirow{2}{*}{\textbf{\highlight{green!50}{9.92}}} &
  \multirow{2}{*}{10.37} &
  \multirow{2}{*}{10.30}
   \\
 &
   &
   &
   &
   &
   &
   &
   &
   &
   &
   &
   &
   &
   &
   &
   \\ \hline
   \multirow{4}{*}{Bengali} &
  \multirow{2}{*}{WER} &
  \multirow{2}{*}{37.70} &
  \multirow{2}{*}{35.44} &
  \multirow{2}{*}{35.90} &
  \multirow{2}{*}{35.09} &
  \multirow{2}{*}{29.65} &
  \multirow{2}{*}{28.77} &
  \multirow{2}{*}{38.10} &
  \multirow{2}{*}{35.60} &
  \multirow{2}{*}{29.96} &
  \multirow{2}{*}{\textbf{\highlight{green!50}{28.50}}} &
  \multirow{2}{*}{\underline{29.30}} &
  \multirow{2}{*}{31.92} &
  \multirow{2}{*}{28.12} &
  \multirow{2}{*}{27.62} \\
 &
   &
   &
   &
   &
   &
   &
   &
   &
   &
   & &
   &
   &
   &
   \\ \cline{2-16}
 &
  \multirow{2}{*}{CER} &
  \multirow{2}{*}{18.44} &
  \multirow{2}{*}{17.73} &
  \multirow{2}{*}{17.08} &
  \multirow{2}{*}{17.22} &
  \multirow{2}{*}{12.76} &
  \multirow{2}{*}{12.58} &
  \multirow{2}{*}{18.59} &
  \multirow{2}{*}{17.72} &
  \multirow{2}{*}{13.06} &
  \multirow{2}{*}{\textbf{\highlight{green!50}{12.38}}} &
  \multirow{2}{*}{\underline{12.52}} &
  \multirow{2}{*}{14.63} &
  \multirow{2}{*}{12.12} &
  \multirow{2}{*}{11.91}
   \\
 &
   &
   &
   &
   &
   &
   &
   &
   &
   &
   &
   & 
   &
   &
   &
   \\ \hline
\end{tabular}%

\end{sc}
    }
    \vspace{3pt}
\caption{\textbf{Parameter-efficient Adaptation Results:} Comparison of WER (\%) and CER (\%) between different parameter-efficient adaptation methods for SeamlessM4T. System A refers to pure ASR fine-tuning, while system T-A refers to text-only adaptation followed by ASR fine-tuning. The best results for System A are \underline{underlined} while the best results for System T-A are in \textbf{bold} for every language. The overall best results have been \highlight{green!50}{highlighted}. }
\label{tabmain}
\vspace {-12pt}
\end{table*}

\subsection{System A: Pure ASR Fine-tuning}

We use the name \textit{System A} to refer to the standard speech-to-text fine-tuning of SeamlessM4T using labeled speech and the ASR objective. The results of this experimental setup are summarized in Table~\ref{table system A-1}. From the results, it is evident that fine-tuning the length adapter requires fewer parameters while providing similar benefits to text decoder fine-tuning across both metrics. Additionally, the ASR fine-tuning of the speech encoder proves to be significantly beneficial, although it involves training a substantially larger number of parameters.


In order to reduce the computational and storage requirements, 
 the fine-tuning was substituted with language-specific adaptations, wherein adapters were introduced in the encoder and decoder, and these were fine-tuned in various combinations using transcribed speech data while freezing the base model. Table \ref{tabmain} depicts the results for the adaptations on System A. The results demonstrate that larger encoder adapters with 50M parameters are the most beneficial in enhancing the ASR performance, achieving WER and CER close to full fine-tuning of the model and the adapters while reducing trainable parameters by $90\%$. Additionally, Table \ref{tabmain} indicates that for the same number of trainable parameters, speech-based training of encoder adapters performs much better than that of decoder adapters. The performance of the length adapter fine-tuning surpasses that of the decoder adapters but falls short compared to the encoder adapters.

\subsection{System T-A: Using Text-only Adaptation}
\vspace {2pt}
The parallel English-target language text data generated by translating the transcripts of IndicVoices data was used to fine-tune the decoder adapters on an English-to-target language MT objective. Table \ref{table system B-1} shows the ASR word error rates (WERs) with the complete transcription data and a smaller 5-hour text data subset (described in Section~\ref{sec:dataset}) to check the comparative benefits of text-only adaptation, without any ASR fine-tuning. For most languages, using the larger text corpus led to better performance. However, the smaller parallel dataset, with significantly fewer tokens, demonstrated comparable performance to that of the complete corpus. This suggests that text-only adaptation can be effective for multilingual multimodal models, even with very limited amounts of data.
\newpage
Moreover, text-only adaptation can be combined with ASR fine-tuning using labeled speech. We refer to the resulting ASR system with text-only adaptation, followed by ASR fine-tuning, as \textit{System T-A}.
Table \ref{tabmain} shows our overall results comparing System A and System T-A. We observe that text-only adaptation followed by ASR fine-tuning is more beneficial than pure ASR fine-tuning, as in System A. The trends of System T-A matched those of System A, with the larger encoder adaptation showing the best performance across all languages except Bengali, the only high-resource language in our study. This suggests that for low-resource languages with limited text and speech data, the most effective strategy is to first use text-only decoder adaptation, followed by speech-based encoder adaptation. It must also be noted that the results of using this strategy are comparable to those after full ASR fine-tuning of the entire model, with a $>90\%$ reduction in the number of trainable parameters, from 571M to 50M.
 \vspace{-5pt}
\begin{table*}[t]
    \footnotesize
    \centering
    \resizebox{\textwidth}{!}{%
    \renewcommand{\arraystretch}{1.2}
\begin{sc}
    \begin{tabular}{l|l|l|l|l|l|l|l}
    
    \hline
       Language 1 & Language 2 & Genetic  &Text- only & \multirow{1}{*}{ASR fine-tuned} & \multirow{1}{*}{Number of}& \multirow{2}{*}{WER}& \multirow{2}{*}{CER} \\
       (Target) & (ASR Fine-tuning) & Distance & Adaptation & Component & Parameters &  \\ \hline 
    \multirow{ 6}{*}{Maithili} &  None & -& No &None&-&82.2&43.39 
    \\\cline{2-8}
    &\multirow{ 4}{*}{Bengali} & \multirow{ 4}{*}{0.625} & No &Length Adapter& 46M & 79.77&40.04 \\
    & & &No& Encoder Adapter & 50M & 81.81&41.61 \\
    & & & No& Len. + Enc. Adapter &52M&80.81&40.44 \\
    & & & Yes& Length Adapter &6M+46M&\textbf{72.52 }
    &\textbf{39.31}\\
    \cline{2-8}
    & \multirow{ 2}{*}{Kannada} & \multirow{ 2}{*}{1.000} & No &Length Adapter&46M&80.29&38.37 \\
    & & & No &Encoder Adapter&50M&85.25&41.58 \\
    \hline
      \multirow{ 6}{*}{Odia} &  None & - & No &None&-&42.81&17.38 \\\cline{2-8}
      & \multirow{ 4}{*}{Bengali} & \multirow{ 4}{*}{0.375} & No &Length adapter& 46M & 41.05&15.07 \\
    & & &No& Encoder Adapter & 50M & 43.67&16.03 \\
    & & & No& Len. + Enc. Adapter &52M&42.4&15.27 \\
    & & & Yes & Length Adapter &6M+46M&\textbf{35.45}
    &\textbf{13.92}\\
    \cline{2-8}
    & \multirow{ 2}{*}{Kannada} & \multirow{ 2}{*}{1.000} & No &Length Adapter&46M&41.21&14.08 \\
    &  & & No &Encoder Adapter&50M&44.01&14.59 \\
    \hline
    \end{tabular}
    \end{sc}}
    \caption{ \textbf{Results for cross-lingual transfer via ASR adaptation:} Comparison of WER(\%) and CER(\%) on low-resource languages with cross-lingual transfer through ASR adaptation of SeamlessM4T. The genetic distances between the (language 1, language 2) pairs suggest that Bengali is related to both the target languages; Kannada, despite being an Indic language, is genetically unrelated to both Maithili and Odia.}
    \vspace {-6pt}
    \label{table:crosslingual}
\end{table*}
\vspace {-1pt}
\subsection{Cross-lingual Transfer}
\vspace {-1pt}
 We hypothesize that the length adapter could capture content-agnostic prosodic characteristics of a language without overfitting on its syntax. Consequently, fine-tuning this adapter using data from a closely related high-resource language might enhance the model's predictions for a low-resource target language. The target languages chosen for this experiment were Maithili and Odia, categorized as zero-shot and low-resource languages for SeamlessM4T, respectively. Bengali, a language belonging to the same Eastern Indo-Aryan language family~\cite{ethnologue2020} as Maithili and Odia, was selected as the high-resource \textit{pivot}. To further justify our choice of the pivot, we examined the genetic distance between the pivot and target languages using lang2vec~\cite{malaviya17emnlp}. Genetic distance \cite{bjerva2019languagerepresentationsreallyrepresent} refers to the measure of divergence between languages based on their evolutionary relationship. The results showed that Bengali was quantifiably close to both target languages. The labeled Bengali speech was used to fine-tune the length adapter and encoder adapters individually and in combination. Separately, Kannada speech was used for length adapter fine-tuning to check if any benefits are obtained with an unrelated language. We also combined this with the text-only adaptation of target language text data to check if both approaches complement each other. Table~\ref{table:crosslingual} summarizes the performance of the cross-lingual systems with both the target low-resource languages. Length adapter fine-tuning outperforms encoder adaptation for cross-lingual transfer. 

 Additionally, we obtained an overall $17\%$ reduction in relative WER for Odia, compared to the base model, by inserting decoder adapters fine-tuned on target language text data into the model whose length adapter was fine-tuned on Bengali ASR data.  
 Thus, for low-resource languages without any speech data, ASR performance may be boosted by length adapter fine-tuning with a closely related pivot language coupled with text adaptation.
\vspace{3pt}
\section{Discussion}


We observe that for decoder adapters, it is more beneficial to use text-only adaptation compared to ASR-based training; the latter's benefit is mainly derived via the encoder layers. This emphasizes the role played by text data in improving the decoder's ability to enhance the internal language model of the ASR system. We also observed that 5-hour text data adaptation, having on average $92\%$ fewer tokens than the full text, performed comparably to full-text data adaptation. This indicates that even limited amounts of text data can significantly boost ASR. 

For a given target language with labeled speech, we found that fine-tuning the encoder adapters was the most accurate and parameter-efficient strategy. However, for cross-lingual zero-shot settings with no labeled data in a target language, we found it beneficial to fine-tune the length adapter with data in a related language rather than fine-tuning encoder adapters; the latter led to overfitting to the related language rather than enabling transfer to the target language. Text-based adaptation led to further improvements in the cross-lingual setting, indicating that even without speech data, ASR for low-resource languages can be improved by fine-tuning the length adapter. Lastly, a curious observation was that higher cross-lingual transfer was seen for genetically closer language pairs, with Odia-Bengali outperforming Maithili-Bengali in terms of relative WER reduction.


\vspace{-2pt}
\section{Conclusion}
\vspace{-2pt}
In this work, we explored the combination of parameter-efficient ASR fine-tuning and text-only adaptation techniques to enhance ASR for low-resource Indic languages using a multi-lingual multi-modal base model (SeamlessM4T). We find that a limited amount of text data was sufficient for adaptation, text-based adaptation was superior to ASR fine-tuning of decoder adapters, and encoder adapters were most effective in limited speech settings. In cross-lingual settings, however, the length adapter (and not the encoder adapter) was most successful, and text adaptation was additionally beneficial. Future work will focus on developing a better understanding of the interplay between different adapters within multimodal models.

\section{Acknowledgements}
The last author would like to gratefully acknowledge the support of the Amazon IITB AI ML Initiative and the consortium project on ``Speech Technologies in Indian Languages" under National Language Translation Mission (NLTM), MeitY, Government of India.


\bibliography{custom}
\bibliographystyle{acl_natbib}
\end{document}